\documentclass{article}

\usepackage{times}
\usepackage{helvet}
\usepackage{courier}
\usepackage{amsmath,amssymb, amsthm} 
\usepackage{graphicx, subfigure}
\usepackage{amssymb}
\usepackage{epstopdf}
\usepackage{enumerate}
\usepackage{array}
\usepackage{bbm}
\usepackage{epstopdf}
\usepackage{subfigure}

\usepackage{natbib}

\usepackage{algorithm}
\usepackage{algorithmic}

\usepackage{hyperref}

\usepackage[accepted]{icml2015arxiv} 

\icmltitlerunning{Dependent Matern Processes}

\begin{document}

\twocolumn[
\icmltitle{Dependent Mat\'ern Processes for Multivariate Time Series}

\icmlauthor{Alexander Vandenberg-Rodes}{vandenbe@uci.edu}
\icmladdress{Department of Statistics, University of California, Irvine}
\icmlauthor{Babak Shahbaba}{babaks@uci.edu}
\icmladdress{Department of Statistics, University of California, Irvine}

\icmlkeywords{Gaussian process}

\vskip 0.3in
]

\begin{abstract}
For the challenging task of modeling multivariate time series, we propose a new class of models that use dependent Mat\'ern processes to capture the underlying structure of data, explain their interdependencies, and predict their unknown values. Although similar models have been proposed in the econometric, statistics, and machine learning literature, our approach has several advantages that distinguish it from existing methods: 1) it is flexible to provide high prediction accuracy, yet its complexity is controlled to avoid overfitting; 2) its interpretability separates it from black-box methods; 3) finally, its computational efficiency makes it scalable for high-dimensional time series. In this paper, we use several simulated and real data sets to illustrate these advantages. We will also briefly discuss some extensions of our model.
\end{abstract}

\newcommand{\E}{\mathbb{E}}
\newcommand{\N}{\mathcal{N}}
\newcommand{\R}{\mathbb{R}}
\newcommand{\C}{\mathbb{C}}
\newcommand{\Z}{\mathbb{Z}}
\renewcommand{\H}{\mathbb{H}}
\renewcommand{\Im}{\operatorname{Im}}
\newcommand{\G}{\mathfrak{G}}
\newcommand{\m}{\mathcal }
\newcommand{\bb}{\mathbb }
\newcommand{\x}{{\bf x}}
\newcommand{\y}{{\bf y}}
\newcommand{\z}{{\bf z}}
\renewcommand{\a}{{\bf a}}
\renewcommand{\b}{\mathbf}
\renewcommand{\c}{{\bf c}}
\renewcommand{\d}{{\bf d}}
\newcommand{\Var}{\operatorname{Var}}
\newcommand{\Cov}{\operatorname{Cov}}
\newtheorem{theorem}{Theorem}
\newtheorem{prop}{Proposition}
\newtheorem{lemma}{Lemma}
\newtheorem{cor}{Corollary}
\newtheorem{defn}{Definition}
\newtheorem{remark}{Remark}
\newtheorem{example}{Example}
\newtheorem*{question}{Question}

\section{Introduction}
Developing powerful models that can capture the dynamics of multivariate time series data, in order to explain their dependencies and predict their unknown values, remains a difficult task in statistics and machine learning. A key challenge is to answer:
\begin{question} How can we describe correlations {\bf among} multiple time series 
\begin{equation} x_1(t), x_2(t), \dotsc, x_p(t),\end{equation} in a way that is also useful for prediction?
\end{question}

In this paper, we tackle this issue by proposing a special case of multivariate Gaussian processes that we call Dependent Mat\'ern Processes (DMP). Similar models have been previously proposed in the econometrics, statistics, and machine learning literature. Here, we follow the recent work of \cite{sarkka13} in considering Gaussian processes from the viewpoint of stochastic differential equations, and attempt to elucidate the mathematical underpinnings of this approach. Despite the similarity to several existing methods, our focus is on constructing a more interpretable model that can explain dependencies among multiple time series, but without sacrificing flexibility or scalability.

This paper is organized as follows. We discuss univariate Gaussian process (GP) models in Section \ref{prelim}, and briefly review several methods to generate multivariate GP. In Section \ref{method}, we present our proposed method, Dependent Mat\'ern Processes. This is a special case of multivariate GP with properties that make it a powerful alternative to existing methods. Section \ref{sde} shows how univariate GPs can be in fact presented either as infinite dimensional functions or as solutions to a specific class of stochastic differential equations. Following \cite{sarkka13}, we show how these two alternative representations are connected, and use this insight to develop the inferential framework of our DMP model in Section \ref{sec:inference}. In Section \ref{experiments}, we present several experiments to illustrate the advantages of our method. Finally, in Section \ref{discussion}, we discuss possible extensions of this approach.

\section{Preliminaries}\label{prelim}
Throughout this paper we will stay within the framework of Gaussian processes (GP). In this section, we discuss univariate and multivariate GP. We represent scalar quantities with lower-case and use capital letters to represent vectors. Boldface capital letters represent matrices. 

\subsection{Univariate Gaussian Processes}\label{sec:univariateGP}
A Gaussian process (GP) on the real line is a random real-valued function $x(t)$, with statistics completely determined by its mean function $\E x(s)$ and {\itshape kernel} $\kappa(s,t) = \Cov(x(s),x(t))$. More precisely, all finite-dimensional distributions $(x(t_1),\dotsc, x(t_n))$ are multivariate Gaussian with mean $(\E x(t_1),\dotsc,\E x(t_n))$, and with covariance matrix $(\kappa(t_k, t_\ell))_{k,\ell = 1}^n$. Since the latter must be positive semi-definite for every finite collection of inputs $t_1,\dotsc, t_n$, only certain kernels $\kappa$ are {\itshape valid} -- that is, define Gaussian processes. Thus when using Gaussian processes, a practitioner often chooses from among the few popular classes of kernels, such as the Squared Exponential (SE), Ornstein-Uhlenbeck (OU), Mat\'ern, Polynomial, and linear combinations of these.

In general, the choice of kernel encodes our {\itshape qualitative} beliefs about the underlying signal. For instance, the OU kernel produces non-differentiable functions $x(t)$, while the SE kernel is infinitely differentiable.
In this paper we will concentrate on the Mat\'ern class of kernels, which have as hyper-parameters the smoothness $\nu$, length-scale $\ell$, and variance $\sigma^2$. In particular, for $n=0,1,\dotsc$ and $\nu = \frac12 + n$ it is known that realizations $x(t)$ of a GP with Mat\'ern kernel are $n$ times continously differentiable, while $\kappa(s,t)$ decays at rate $e^{-|t-s|\sqrt{2\nu}/\ell}$ as $t-s$ becomes large~\cite{stein99, rasmussen06}. It is then straightforward to add extra {\itshape observation noise} reflecting our uncertainty in the accuracy of our measurements.

\subsection{Multivariate GPs}
There often arise situations where we would like to jointly model several time series $x_1(t), x_2(t), \dotsc, x_p(t)$, for the purpose of {\itshape inference}, in particular attempting to quantify the dependencies between the observed series, and/or to improve our {\itshape predictions} of one series using data from the others. In the context of Gaussian processes, we intend that for different processes $i$ and $j$ we have a non-zero covariance. In fact, a {\itshape multi-output} or {\itshape multivariate} Gaussian process can be defined just as in Section \ref{sec:univariateGP}, but where the kernel function now depends on two pairs of inputs. For simplicity we will assume in what follows that the mean of each time series is the zero function. The kernel $\kappa$ is now defined for $i,j=1,\dotsc, p$ and $s,t\in \R$ as
\begin{equation}\label{full_cov}
\kappa([i,s], [j,t]) = \E x_i(s)x_j(t).
\end{equation}
The initial challenge within the Gaussian process context is to produce a valid and interpretable kernel. 

\subsubsection{Linear models}
The usual technique for generating multivariate GP kernels is known as {\itshape co-kriging} from the geostatistical literature~\cite{cressie93}. The simplest case is known as the {\itshape intrinsic co-regionalization model} (ICM), where one takes $\b C = (c_{ij})$ to be a positive definite $p\times p$ matrix, $\kappa^{(1)}(s,t)$ to be a valid univariate kernel, and defines the multi-output kernel $\kappa$ to be their product
\begin{equation}
\E x_i(s)x_j(t) = c_{ij}\kappa^{(1)}(s,t).
\end{equation}
Notice that while the intrinsic co-regionalization model is easily interpretable -- the single matrix $\b C$ provides the covariances between the time series -- all outputs share the same univariate kernel, which makes for a rather inflexible model.

The {\itshape linear model of coregionalization} (LCM) adds more flexibility by allowing linear combinations of ICM's, resulting in a kernel of the form

\begin{equation}\label{lcm}
\E x_i(s)x_j(t) = \sum_{k=1}^q c_{ij}^{(k)} \kappa^{(k)}(s,t).
\end{equation}
For each $k=1,\dotsc, q$, $\kappa^{(k)}(s,t)$ is assumed to be a valid kernel for a univariate GP, and $\b C^{(k)} = (c^{(k)}_{ij})$ is assumed to be a positive definite matrix. It is not hard to see that \eqref{lcm} results in a valid kernel. However, we have now lost some the interpretability of the ICM. More problematically, the LCM still does not provide a notion of correlation between processes with differing length-scales. One proposed solution is the {\itshape process convolution} approach~\cite{boyle05, alvarez2011conv}, which allows for qualitatively very different processes to be correlated, though with some loss of interpretability.

\subsubsection{Latent models}
Another approach is to describe $(x_1(t), \dotsc, x_p(t))$ as linear combinations of {\itshape latent} factors. We suppose $u_1(t), \dotsc, u_q(t)$ are independent mean zero Gaussian processes, and let
\begin{equation}\label{latent}
x_i(t) = \sum_{k=1}^q a_{i,k} u_k(t), \quad \text{ for } i=1,2,\dotsc p.
\end{equation}

Let $\kappa_i(s,t)$ = $\E u_i(s) u_i(t)$ be the kernel for the $i$'th latent process. Then the observed processes $\mathbf{x}(t) = (x_1(t),\dotsc, x_p(t))$ are jointly mean-zero Gaussian with covariances 
\begin{equation}\label{LMC}
\E x_i(s) x_j(t) = \sum_{k=1}^q a_{i,k}a_{j,k} \kappa_k(s,t).
\end{equation}
 This is the {\itshape semi-parametric latent factor model} of \cite{teh05}, so-called because the linear combination of latent GP's is parameterized by the matrix of coefficients $A = (a_{i,k})$, while each Gaussian process is of course a non-parametric model. However, this latent model \eqref{latent} is actually an example of the above linear model of coregionalization, where $\b C^{(k)}$ is just the outer product of the vector $a_{\cdot, k}$ with itself.

See \cite{alvarez11} for a nice survey of these and other variants of co-kriging used in the machine learning literature.

\subsubsection{Other approaches}
Instead of trying to create multivariate kernels in a general fashion, one can attempt multivariate generalizations of a given class of univarate kernels, often by using B\^ochner's theorem (see Section \ref{sec:infdim} below). The recent work of \cite{gneiting10, apanasovich12} is perhaps the most relevant to our model, as they show how to construct a family of valid kernels for multivariate Gaussian processes on $\R^d$ where the marginal processes each have Mat\'ern kernel with different hyperparameters.

\section{Dependent Mat\'ern processes}\label{method}
From a modeling perspective we would like to describe correlations between processes that have different (unique) hyperparameters, whereas in most of the above models this is only roughly attained by taking linear combinations of processes.

\subsection{Our approach} We will model multivariate time series $X(t) = (x_1(t), \dotsc, x_p(t))$ such that each marginal process $x_i(t)$ is a stationary mean-zero Gaussian process with Mat\'ern kernel
\[\kappa_{\nu, \ell_j, \sigma_j}(t) = \E x_j(0)x_j(t),\] thus the processes are allowed different length scales and variance, while sharing a common smoothness.  In what follows we will always assume $n = \nu-\frac12$ to be an integer. As we will explain in Section \ref{sec:connection}, each $x_j(t)$ can actually be represented as a solution of the stochastic differential equation
\begin{equation}\label{matsde}
\left(\frac d{dt} + \frac{\sqrt{2\nu}}{\ell_j}\right)^{n+1} x_j(t) = \sigma_j C_{\nu, \ell_j}\dot w_j(t),
\end{equation}
where $\dot w(t)$ is white noise and $C_{\nu,\ell_j}$ is a constant.
\subsubsection{A new multi-output GP}
To introduce dependence among the Mat\'ern processes $x_j(t)$ we {\itshape correlate the input noises} $\sigma_j\dot w_j(t)$ in \eqref{matsde}. That is, we let $\b L$ be a $p\times R$ matrix and set 
\begin{equation}\label{sharednoise}
(w_1(t),\dotsc, w_p(t))^T = diag(\sigma_1^{-1},\dotsc, \sigma_p^{-1})\b L V(t),
\end{equation} where $V(t)$ is a vector of $R$ independent standard Brownian motions, which we can think of as latent noise processes. Note that $\b L$ has absorbed the $\sigma_j$ parameters, and $(c_{ij}) = \b C=\b L\b L^T$ is the covariance matrix of the input noises.

The stationary solution of these coupled SDEs results in multi-output GP, which we will refer to as a {\itshape Dependent Mat\'ern process}.

 In Section \ref{sec:inference} we will show how to compute the kernel \eqref{full_cov} for this new process, resulting in (for $\nu=\frac12$) 
\begin{equation}\label{cov_mat12}
\E x_i(s)x_j(t) \propto c_{ij}r_{ij}e^{-(t-s)/\ell_j},
\end{equation}
while for $\nu=\frac32$ we obtain
\begin{equation}\label{cov_mat32}
\E x_i(s)x_j(t) \propto  c_{ij}r_{ij}^3\frac{2+(t-s)\left(\frac{\sqrt{3}}{\ell_i} + \frac{\sqrt{3}}{\ell_j}\right)}{e^{\sqrt3(t-s)/\ell_j}}.
\end{equation}
In both cases we are assuming $s\leq t$, and the factor 
\begin{equation}\label{rij}
r_{ij} = 2\sqrt{\ell_i \ell_j}/(\ell_i + \ell_j)
\end{equation} is the ratio of the geometric and arithmetic means of the two length-scales.

Examining these two expressions for the kernel, one should note:
\begin{enumerate}
\item They are not symmetric in time, as interchanging $s$ and $t$ would replace $\ell_j$ with $\ell_i$ in the exponential. That is, the covariance kernel respects the forward flow of time, which we believe to be a desired characteristic. Note this feature is missing from all of the models discussed above.

\item For $\ell_i\approx\ell_j$ the $r_{ij}$ factor is close to $1$, but as $\ell_i$ and $\ell_j$ increasingly differ in scale $r_{ij}$ goes to zero. Intuitively this means that two processes with different length scales cannot move tightly together.
\end{enumerate}
\subsubsection{Defining the correlation}

Even if the various length scales are quite different, the matrix $\b C = (c_{ij})$, which we can recover from observed data, can be normalized in the usual way to obtain a clear, though model-dependent, notion of {\itshape correlation $(\rho_{ij})$ between time series}:
\begin{scriptsize}
\begin{equation}
\begin{pmatrix}
\rho_{11} & \cdots &\rho_{p1} \\
\vdots & \ddots & \\
\rho_{p1} & \cdots & \rho_{pp}
\end{pmatrix} = \begin{pmatrix}
c_{11}^{-\frac12} &&\\
& \ddots & \\
&& c_{pp}^{-\frac12}
\end{pmatrix} C \begin{pmatrix}
c_{11}^{-\frac12} &&\\
& \ddots & \\
&& c_{pp}^{-\frac12}
\end{pmatrix}.
\end{equation}
\end{scriptsize}

\subsubsection{Latent force models}\label{sec:latentforce}
Our proposed model can be thought of as a particular case of {\itshape latent force models}~\cite{alvarez09}, although our motivation and approach to inference are very different. With a latent force model one thinks of each output time series as following specific physical dynamics, such as a damped harmonic oscillator, but that is also under the influence of latent forces (modelled as GPs), which are shared across the outputs as we do in \eqref{sharednoise}.

With our model we are more interested in providing a {\itshape interpretable} notion of correlation between the time series, and do not assume knowledge of any underlying physical dynamics for the outputs. We are instead interested in the qualitative features of the Mat\'ern class, and following, e.g. \cite{hartikainen10, mbalawata13, sarkka13} we construct the SDE dynamics that represent such processes.

\subsection{Computational complexity}
Gaussian processes in general suffer from the {\itshape big-N} problem, that is, computations involving a $N$ samples from a Gaussian process typically are of cubic complexity in $N$, since one usually needs to invert the $N\times N$ covariance matrix $(\kappa(t_i, t_j))$. In the case of $p$ processes sampled $N$ times, the resulting computational cost is $O(N^3p^3)$, which can be already prohibitive when there are only a few hundred samples.

In special cases such as equally-spaced observations there are faster techniques such as circulant embedding~\cite{dietrich97}, and for general Gaussian processes there has been a recent flurry of research into sparse approximations~\cite{quinonero05}.

As we will see in Section \ref{sec:stationary}, stochastic differential equations can be transformed into state space models, which have the nice feature that computing the likelihood of $N$ observations, or using the Kalman filter and Rauch-Tung-Streibel smoother for prediction, only has complexity $O(p^3N)$. This allows our model to easily handle data containing thousands of observations.

In order to transform our DMP model into a state space model we first need to set up the mathematical background connecting Gaussian processes and stochastic differential equations. However, the reader might prefer to jump to Section \ref{sec:kalman}, where we show how to use the state space form for inference and prediction.

\section{Two approaches to univariate Gaussian processes}\label{sde}

\subsection{Infinite dimensional regression}\label{sec:infdim}
One way of viewing a Gaussian process is as a random function of the form
\begin{equation}\label{gpfunc}
x(t) = \sum_{k=0}^\infty a_k \psi_k(t),
\end{equation}
where $\{\psi_k(t)\}$ is a collection of deterministic square integrable ($L^2$) functions, that is, {\itshape features}, and we place an iid $N(0, 1)$ prior on the coefficients $a_k$. As a linear combination of Gaussians is Gaussian, $x(t)$ is clearly a GP. 

To compute the kernel of \eqref{gpfunc}, we take {\itshape any} orthonormal basis $\{\phi_n(t)\}$ of $L^2$, let $g$ be an integrable function, and define $\psi_k(t)$ to be the convolution $\int \phi_k(u)g(t-u) du$, which should be thought of as the $L^2$ inner product of $\phi_k$ and $g(t-\cdot)$. Since $\E a_na_m = \delta_{n,m}$, the kernel $\E x(s)x(t)$ reduces to
\begin{equation}\label{inner}
\sum_{k=0}^\infty \psi_k(s)\psi_k(t) = \int g(s-u) g(t-u) du.
\end{equation}
The last equality is just Parsival's identity, relating the inner product of $g(s-\cdot)$ and $g(t-\cdot)$ to the dot product of their coefficient vectors $\{\psi_k(s)\}$ and $\{\psi_k(t)\}$ in the orthonormal basis. The key point of \eqref{inner} -- similar to the {\itshape kernel trick} for support vector machines -- is that this kernel is independent of the choice of orthonormal basis, and so the function $g$ now defines the Gaussian process.

By using the Fourier transform we can characterize the class of valid kernels. Given a {\itshape stationary} kernel ($\kappa(s,t) = \kappa(0, t-s)$), its {\itshape spectral density} $S(\xi)$ is defined by:
\begin{equation}\label{spectral density}
\kappa(0,t) = \int e^{it\xi} S(\xi) d\xi.
\end{equation}
Noting that the right hand side of \eqref{inner} describes a stationary kernel, we use Parsival's identity again to see that 
\begin{equation}\label{hatg}
\int g(t-u)g(-u) = \int e^{it\xi} \lvert \hat g(\xi)\rvert^2 d\xi,
\end{equation}
with $\hat g$ the Fourier transform of $g$. In particular, a function $\kappa(t)$ with non-negative Fourier transform (spectral density) is a valid kernel for a stationary GP; the precise equivalence, known as B\^ochner's theorem~\cite{stein99}, shows that all valid kernels arise in this fashion.

\subsection{Stochastic differential equations} A particularly nice way to construct Gaussian processes on the real line is via solutions of stochastic differential equations (SDE's). Although constructing Gaussian processes through SDE's goes back to the seminal article of~\cite{doob44}, and has been used extensively in econometrics~\cite{bergstrom90}, it has only recently seen development in the machine learning literature \cite{hartikainen10, hartikainen12, mbalawata13, sarkka13, reece14}.

The archtypical SDE is the Ornstein-Uhlenbeck process
\begin{equation}\label{OUsde}
\frac{dx}{dt} (t) = \alpha x(t) + \dot w(t),
\end{equation}
where $\dot w(t)$ is Gaussian white noise. One can make mathematical sense of this via its integrated form
\begin{equation}\label{OUint}
x(t)-x(s) = \int_s^t \alpha x(u) du + w(t)-w(s),
\end{equation}
with $w(t)$ as the Weiner process (Brownian motion). Given an initial value $x(s)$, it has the solution 
\begin{equation}\label{stochint}
x(t) = e^{(t-s)\alpha}x(s) + \int_s^t e^{(t-u)\alpha}dw(u), \quad t\geq s.
\end{equation}
Although it is sometimes thought that making sense of the integral in \eqref{stochint} requires the full weight of It\^{o} calculus, for deterministic (and differentiable) integrands we can use the integration by parts formula: $\int_s^t f(u) dw(u) = w(t)f(t) - w(s)f(s) - \int_s^t f'(u)w(u) du$.\footnote{In this interpretation only an interchange of integrals is required to show that \eqref{stochint} solves \eqref{OUint}.} 

\subsubsection{General case}
Higher order SDE's of the form
\begin{equation}\label{thesde}
\frac{d^n}{dt^n}x(t)+ a_{n-1}\frac{d^{n-1}}{dt^{n-1}}x(t) + \cdots + a_0 x(t) = \sigma\dot w(t),
\end{equation}
such as the one defining the Mat\'ern process \eqref{matsde}, are similarly interpreted. Letting $F(t)$ be the vector of derivatives $(x(t), x'(t), \dotsc, x^{(n-1)}(t))^T$, we can rewrite \eqref{thesde} as
\begin{equation}\label{sde2}
\frac{dF}{dt}(t) = \begin{pmatrix}
0 & 1 & &\\
 & \ddots & \ddots & \\
 & & 0 & 1\\
 -a_0 & -a_1 & \cdots & -a_{n-1}
\end{pmatrix} F(t) + \begin{pmatrix} 0 \\ \vdots \\ 0 \\ \sigma\end{pmatrix} \dot w(t).
\end{equation}
With $\b Q$ as the $n\times n$ matrix above and $E = (0,\dotsc, 0,\sigma)^T$, the solution to \eqref{thesde} is completely analogous to \eqref{stochint}:
\begin{equation}\label{stsol}
F(t) = e^{(t-s)\b Q} F(s) + \int_s^t e^{(t-u)\b Q} E dw(u).
\end{equation}

\subsubsection{Stationarity}\label{sec:stationary} We now require that the eigenvalues of $\b Q$, that is, the zeros of its characteristic polynomial 
\begin{equation}\label{poly}
x^n + a_{n-1} x^{n-1} + \cdots + a_0,
\end{equation} all have negative real part. In this case, taking the limit of \eqref{stsol} as $t\rightarrow\infty$ results in a zero mean Gaussian random vector with a finite covariance matrix we denote by $\Sigma_\infty$. If we then choose some initial point $F(0)\sim \mathcal N(0,\Sigma_\infty)$, the resulting process $(F(t);\; t\geq 0)$ is a stationary $n$-dimensional Gaussian Markov process, with covariance kernel 
\begin{equation}\label{cov}
\E F(s) F(t)^T = e^{(t-s)\b Q}\Sigma_\infty.
\end{equation} 
The integral in \eqref{stsol} is also Gaussian with covariance
\begin{equation}\label{processnoise}
\Sigma_\infty - e^{(t-s)\b Q} \Sigma_\infty e^{(t-s)\b Q^T}.
\end{equation}
Usually we only observe the positions $x(t)$ at a finite collection of times $t_1,\dotsc, t_N$. Assuming corruption by observation noise $\epsilon_k$, the resulting observations of the SDE \eqref{thesde} can be written in the following state space form:
\begin{align}\label{statespace}
F(t_k) &= e^{(t_k-t_{k-1})\b Q}F(t_{k-1}) + \eta_k,\\
y(t_k) &= H F(t_k) + \epsilon_k,\label{ss2}
\end{align}
where $\{\eta_k\}$ are independent Gaussian with covariance \eqref{processnoise}, and $H=(1, 0, \dotsc, 0)$ is the observation matrix.

\subsection{Connecting SDEs to GPs}\label{sec:connection}
Unfortunately not all Gaussian processes on $\R$ exactly correspond to an SDE. The precise relationship is due to \cite{doob44}: A stationary Gaussian process on $\R$ can be represented as the stationary solution of \eqref{thesde} when its spectral density has the form
\begin{equation}\label{spectral}
S(\xi) = \frac{\sigma^2}{|(i\xi)^n + a_{n-1}(i\xi)^{n-1}+ \cdots + a_1(i\xi) + a_0 |^2}.
\end{equation}

The fundamental example is the Mat\'ern class of Gaussian processes, which have a kernel with spectral density 
\begin{equation}\label{matspec}
S(\xi) = \sigma^2 C^2_{\nu, \ell} \left(\xi^2 + \frac{2\nu}{\ell^2} \right)^{-(\nu+\frac12)},
\end{equation} where $\nu$, $\ell$, and $\sigma$ are the smoothness, lengthscale, and variance parameters, respectively, and $C_{\nu, \ell}$ is a constant with respect to $\xi$ and $\sigma$. When $\nu = n+\frac12$ we can factor $S(\xi) = \hat g(\xi) \overline{\hat g(\xi)}$, where\footnote{There are multiple choices for $\hat g(\xi)$, however, only this one ensures that the zeros of $1/\hat{g}(ix)$ (that is, the polynomial \eqref{poly}) all have negative real part, and thus corresponds to a stationary SDE as discussed in Section \ref{sec:stationary}.}
\begin{equation}
\hat g(\xi) = \sigma C_{\nu, \ell}\left( i\xi + \frac{\sqrt{2\nu}}{\ell}\right)^{-n-1}.
\end{equation}
Hence such Mat\'ern class GP's can be realized as solutions of the SDE \eqref{matsde} used in our multivariate GP.

In order to put $\eqref{matsde}$ into the state space form \eqref{statespace}, we expand out its left hand side using the binomial theorem to obtain the $n+1\times n+1$ matrix $\b Q$ in \eqref{sde2}. With $n=0$ or $1$ (corresponding to $\nu=1/2$ or $3/2$), we have
\begin{equation}\label{Qmat}
\b Q_j = (1/\ell_j), \mbox{ or } \b Q_j = \begin{pmatrix}
0 & 1\\
-\frac3{\ell_j^2} & -\frac{2\sqrt{3}}{\ell_j}
\end{pmatrix}.
\end{equation}
Furthermore, each matrix exponential can be computed analytically~\cite{jones81}. With $n=1$, for example, we have 
\begin{equation}\label{expm}
 e^{t\b Q_j} = e^{-t\sqrt{3}/\ell_j} \begin{pmatrix} 1+\frac{t\sqrt{3}}{\ell_j} & t\\ -\frac{3t}{\ell_j^2} & 1-\frac{t\sqrt{3}}{\ell_j}\end{pmatrix}.
 \end{equation}
 
Although we will not make use of it here, one should note that by approximating $\hat g$ \eqref{hatg} with rational functions, one can approximately represent other Gaussian processes in terms of SDEs~\cite{sarkka13, solin14}

\section{Inference in the dependent Mat\'ern model}\label{sec:inference}
To obtain the {\itshape joint} state space representation of the $p$ coupled SDE's \eqref{matsde}, we stack the $p$ derivative vectors $F_1, \dotsc, F_p$ together to create the length $p(n+1)$ vector 
\begin{equation}
\vec F(t) = (x_1(t),\dotsc, x_1^{(n)}(t),\dotsc, x_p(t), \dotsc, x_p^{(n)}(t))^T,
\end{equation} containing the $p$ processes and their first $n$ derivatives.

Recalling \eqref{statespace} and \eqref{ss2}, write $\b E = E\otimes \b I_p$, $\b H = H\otimes \b I_p$,  where $\b I_p$ is the $p\times p$ identity matrix and $\b A \otimes \b B$ is the Kronecker product of the matrices $\b A$ and $\b B$. In particular, $\b H \vec F(t) = (x_1(t),\dotsc, x_p(t))$. Then with $\vec{\b Q}$ as the block diagonal matrix with blocks $\b Q_1,\dotsc, \b Q_p$ as in \eqref{Qmat}, the equivalent state space formulation of the coupled SDE's \eqref{matsde} can be written as
 \begin{align}\label{statespace1}
 & \vec F(t_k) = e^{(t_k - t_{k-1})\vec{\b Q}} \vec F(t_{k-1}) + \vec{\eta}_k \\
 & Y(t_k) = \b H \vec F(t_k) + \vec\epsilon_k. \label{statespace2}
 \end{align}
Note that the matrix exponential is just a block diagonal matrix with blocks $e^{-\Delta t_k \b Q_j}$.
The observation noise $\vec\epsilon_k$ is assumed to be iid mean-zero Gaussian with diagonal covariance $diag(\tau^2_1,\dotsc, \tau^2_p)$. And finally the process noise $\vec{\eta}_k$ is given by \eqref{processnoise}. We omit the calculation of the needed stationary covariance $\Sigma_\infty$ of $\vec{F}(t)$, which is a block matrix with the  $i,j$-block an $n+1\times n+1$ matrix
\begin{equation}\label{blk1}
B_{ij} = c_{ij}r_{ij},\quad \mbox{ if }n=0,
\end{equation}
where $r_{ij}$ was defined in \eqref{rij}, and when $n=1$:
\begin{equation}\label{blk2}
B_{ij} = c_{ij}r_{ij}^3\begin{pmatrix}
2 & \frac{\sqrt 3}{\ell_i} - \frac{\sqrt 3}{\ell_j} \\
\frac{\sqrt 3}{\ell_j} - \frac{\sqrt 3}{\ell_i} & \frac{6}{\ell_i\ell_j}
\end{pmatrix}.
\end{equation}

Finally, by substituting \eqref{blk1} and \eqref{blk2} into \eqref{cov}, we can obtain the covariances \eqref{cov_mat12} and \eqref{cov_mat32}.

 \subsection{Applying the Kalman filter and smoother}\label{sec:kalman}
Given a state space model
\begin{align*}
z_k &= A_kz_{k-1} + \eta_k,\\
y_k &= H_k z_k + \epsilon_k,
\end{align*}
and observed data $y_1, \dotsc, y_N$, with $\eta_k$ and $\epsilon_k$ as independent Gaussian noise, the Kalman filter~(see \citet{murphy12} for example) recursively calculates the conditional means and covariances 
\begin{align}
m_k^- =& \ \E (z_k |\ y_1, \dotsc, y_{k-1}, \Theta)\\
P_k^- =& \ \E \left((z_k - m_k^-)(z_k - m_k^-)^T | y_1, \dotsc, y_{k-1}, \Theta\right).
\end{align}
We use $\Theta$ to denote the collected parameters for $A_k, \eta_k$, and $\epsilon_k$.
Setting $S_k = H_k P_k^{-} H_k^T + J_k$, where $J_k$ is the covariance matrix of the observation noise $\epsilon_k$, the log likelihood $\log \bb P(\Theta | y_1,\dotsc, y_N)$ is, up to a constant,
\begin{align}\notag
&\log \bb P(\Theta) + \sum_{k=1}^N \log \bb P(y_k | y_1, \dotsc, y_{k-1}, \Theta)  \\
=&\log \bb P(\Theta) +  \sum_{k=1}^N \log \mathcal N(y_k; H_k m_k^-, S_k). \label{loglike}
\end{align}

For prediction we can use the Rauch-Tung-Streibel smoother to obtain the means and covariances,
\begin{align}\label{RTS1}
m_{k;N} =& \ \E(z_k | y_1,\dotsc, y_N, \Theta),\\
P_{k;N} =& \ \E\left((z_k-m_{k;N})(z_k - m_{k;N})^T | y_1, \dotsc, y_N, \Theta\right),\label{RTS2}
\end{align}
conditional on the training data and the inferred parameters $\Theta$. Note that the state space framework easily handles the missing (test) data by modifying the observation matrix $H_k$. 

\subsection{Implementation} 
In the case of our state space model \eqref{statespace1} and \eqref{statespace2}, we assume that the smoothness $\nu$, and the number of latent noise processes $R$ is chosen ahead of time. Hence our collected parameters $\Theta$ are: $\ell_1,\dotsc, \ell_p$ (length-scale parameters), $\b L$ (a $p\times R$ matrix parameterizing the covariance across the observed processes), and $\tau_1^2,\dotsc, \tau_p^2$ (variances of the observation noise). Our inference involves two stages: 
 \begin{enumerate}
 \item Taking the state space form \eqref{statespace}, \eqref{ss2}, of each {\itshape univariate} Mat\'ern process $x_j(t)$, we estimate the individual length-scales $\ell_j$ one-by-one by minimizing \eqref{loglike}. In practice, we found that Matlab's fminunc() works well.
 \item Now using the state space form \eqref{statespace1}, \eqref{statespace2} for the multi-output process, we sample from the posterior distribution of the remaining parameters $\b L$ and $\vec\tau^2$, using the Metropolis-Hastings algorithm.
 \end{enumerate}

\section{Experiments}\label{experiments}
In this section, we use simulated and real data to evaluate our method. We compare our method to some existing algorithms in terms of prediction accuracy. Additionally, we show how our method describes correlations among multiple time series.  

\begin{figure}[!h]
  \begin{center}
  \includegraphics[height=1.5in, width=3in]{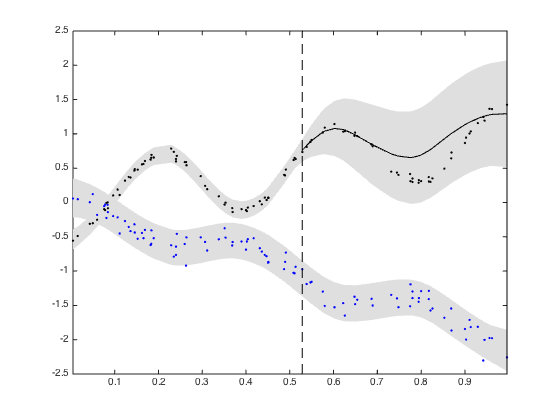}
  \caption{Simulated time series: $x_1(t) =0.2\cos(5\pi t) -2t + 0.1\epsilon$ and $x_2(t) = t - 0.5 \cos(5\pi t)  + 0.04\eta$ for $t \in [0, 1]$. The observed samples are shown with blue and black dots, respectively. The black line illustrates the Kalman-filter predicted means for the withheld samples.}
  \label{sim1}
  \end{center}
\end{figure}

\begin{figure*}[!t]
  \begin{center}
 	\subfigure[$\nu=\frac12$, SMSE: $0.059$]{\includegraphics[height=1.5in, width=2.2in]{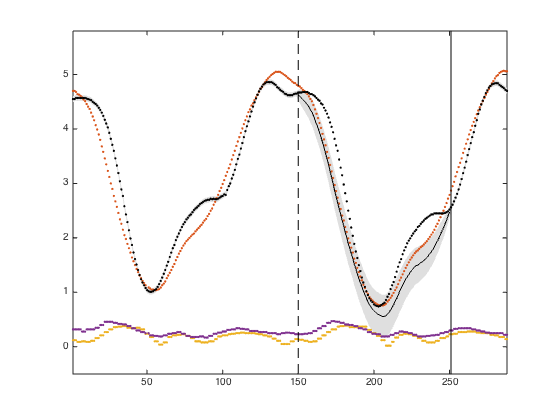}}	
	\subfigure[$\nu=\frac32$, SMSE: $0.022$]{\includegraphics[height=1.5in, width=2.2in]{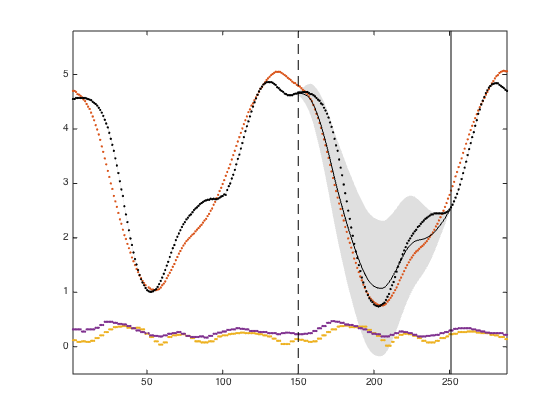}}	
	\subfigure[$\nu=\frac52$, SMSE: $0.224$]{\includegraphics[height=1.5in, width=2.2in]{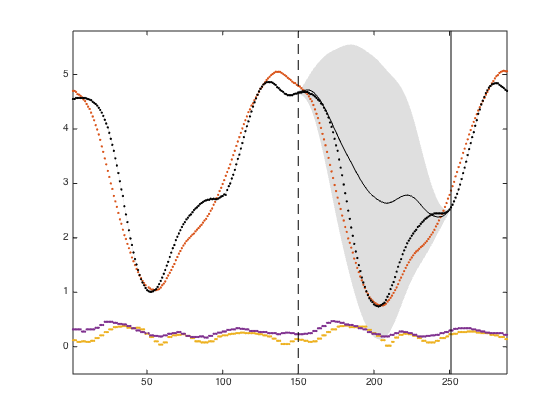}}	
  \caption{Wave and Tide data-- The values of Sotonmet tide heights between the two vertical lines are assumed to be unknown. The black dots represent the true values of the Sotonmet tide heights, and the black lines show the predicted mean, with $\pm 2$ standard deviations shaded, using three choices of smoothness parameters $\nu$ in our model. The standardized mean squared errors (SMSE) are provided for each option.}
  \label{fig:waveht}
  \end{center}
\end{figure*}

\subsection{Synthetic data}\label{sec:synth}
For our first experiment, we took a random selection of 100 times $t \in [0, 1]$, and simulated two time series 
\begin{align*}
x_1(t) =&\ 0.2\cos(5\pi t) -2t + 0.1\epsilon, \\
x_2(t) =&\ t - 0.5 \cos(5\pi t)  + 0.04\eta,
\end{align*}
where $\epsilon$ and $\eta$ are iid $\mathcal N(0,1)$ noises. These are shown as the blue and black dots, respectively, in Figure \ref{sim1}. We removed the last 41 observations of the second time series (black dots), illustrated by the vertical dashed line, and treated them as the test set. The black line shows the Kalman filter predicted means for the withheld data, using the last sampled parameters based on our model, and the gray area shows the given $\pm 2\sigma$ deviations about the predicted mean for both series. On a 2011 Macbook with a 2.3Ghz i5 processor and 8GBs of RAM it took 65 seconds to draw 50,000 posterior samples of the correlation and noise parameters, while estimating the length scales and predicting the missing values is near-instantaneous. A highly optimized Kalman filter routine might lower the sampling time by an order of magnitude.

\subsection{Wave and Tide data}\label{sec:wave-tide}
For our second experiment, we tested our model on wave and tide data from the weather stations of Cambermet, Chimet, and Sotonmet, all on the southern coast of the U.K.\footnote{Data available from \hyperref[http://www.chimet.co.uk]{http://www.chimet.co.uk}}. The data consists of four time series: the tide heights of Chimet and Sotonmet, and the wave heights of Cambermet and Chimet. There are 288 observations, taken at 5 minute intervals, from the day of January 1, 2010. Observations 150 to 250 of the Sotonmet tide heights (black dots) were removed to make a test set.

With this data we investigated how different choices of the smoothness parameter $\nu$ affected performance. All simulations used $R=4$ independent noise sources. In figure \ref{fig:waveht} the black dots represent the true values of the Sotonmet tide heights, and the black line is the predicted mean, with $\pm 2$ standard deviations shaded.

With $\nu=1/2$ the model overestimates the correlation between the two tide heights ($\rho\approx 0.9$), resulting in an overconfident estimate that tracks the other tide height (red dots) too closely. With $\nu=5/2$ we have the opposite problem: despite the two tide heights staying together over the course of the day, there is not much correlation found between their third derivatives, resulting in a very weak prediction. The middle case of $\nu=3/2$ strikes a nice balance, with estimated length-scales of 
$(102.5, 75.4, 24.0, 41.0)$, and inferred correlation matrix
{\begin{scriptsize}
\[\begin{pmatrix}
1.0000  &  0.6155 &   0.0191 &   0.0655 \\
    0.6155 &   1.0000 &   0.0547  &  0.0984\\
    0.0191  &  0.0547  &  1.0000   & 0.3344\\
    0.0655  &  0.0984  &  0.3344  &  1.0000
\end{pmatrix}\]
\end{scriptsize}
}
Note the moderate correlations ($\approx 0.6$) found between the tide heights, and the weak correlations ($<0.1$) between the tide and wave heights.

\begin{figure*}[!t]
  \begin{center}
  \subfigure[CAD]{\includegraphics[height=1.5in, width=2.2in]{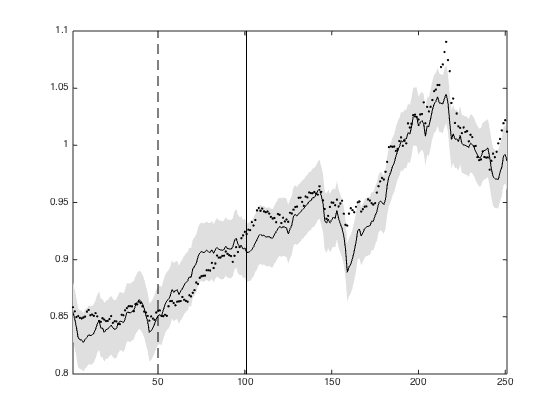}}
  \subfigure[JPY]{\includegraphics[height=1.5in, width=2.2in]{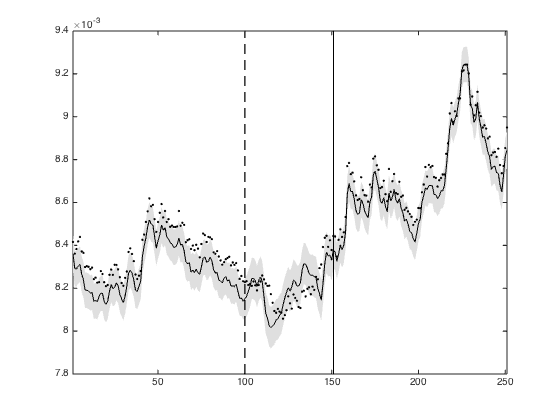}}
  \subfigure[AUD]{\includegraphics[height=1.5in, width=2.2in]{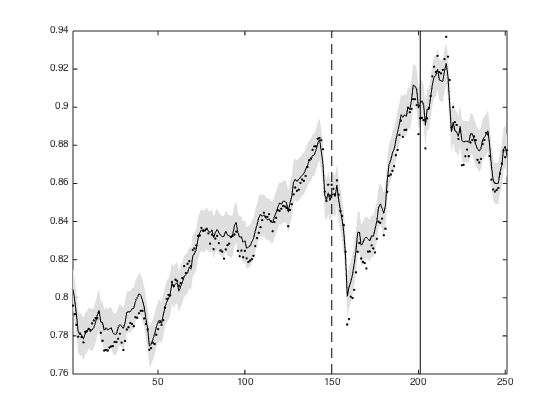}}
  \caption{The US Dollar exchange rate with respect to Canadian Dollar (CAD), Japanese Yen (JPY), and Australian Dollar (AUD). The black dots show the observed data. The vertical lines show the intervals where the data are assumed to be unknown (i.e., test set). The solid lines show the predicted means using our model. The grey areas show the corresponding 95\% intervals.}
  \label{financial}
  \end{center}
\end{figure*}

\subsection{Financial data}
For our last example, we consider the inference of missing data in the multivariate financial dataset used in \cite{alvarez10}. It contains thirteen time-series for the US Dollar exchange rate with respect to the top 10 international currencies (Canadian Dollar, Euro, Japanese Yen, Great British Pound, Swiss Franc, Australian Dollar, Hong Kong Dollar, New Zealand Dollar, South Korean Won, Mexican Peso), and three precious metals (gold, silver, platinum), over all 251 working days of the 2007 calendar year. Following \cite{alvarez10} we removed the mean and normalized each series to have unit variance, and removed a test set of 251 data points, covering days 50-100, 100-150, and 150-200, from the Canadian Dollar, Japanese Yen, and Australian Dollar series, respectively. The remaining 3051 data points were used as the training set. (There are already 59 missing data points from the precious metal series). These three time series are shown in Figure \ref{financial}, along with the predicted means. As before, the vertical lines show the intervals where the test set data was withheld.

Because of the roughness of the paths we modeled the 13 time series as dependent Mat\'ern($\nu=\frac12$) processes, and restricted the parameter space by allowing for only $R=4$ independent noise sources. The predicted means and the corresponding 95\% intervals are shown as solid lines and shaded areas respectively in Figure \ref{financial}. We then compared with the linear model of coregionalization (LMC) where the kernel \eqref{lcm} is a combination of two Mat\'ern($\nu=\frac12$) kernels, and $\b C^{(1)}$ and $\b C^{(2)}$ are both of rank 2. Our model's predictions had a standardized mean squared error (SMSE) of 0.087 (averaged across the three test outputs), while the LMC scored 0.49. Note that in this case our model is essentially the Stochastic Latent Force model in of \cite{alvarez09}, with all four latent processes as white noise. Nonetheless we end up with much better predictions (for their best model with one smooth and three white noise latent processes, \cite{alvarez10} quote a SMSE of 0.2795, and 0.39 for their LMC implementation). We believe this shows the power of independently modelling the output processes and then their correlations. 

\section{Discussion}\label{discussion}
In this paper, we have proposed a new class of stochastic process models for multivariate time series. Using several examples, we illustrated our method's predictive power and interpretability. However, as discussed above, our method is also designed to be extendable to problems with more complex structures.

One possible extension to our model would be to allow kernels with (quasi-)periodic behavior, leading to better inference when modeling periodic phenomena such as the wave and tide data of Section \ref{sec:wave-tide}. This is indeed possible within the state space approach, as exemplifed by the stochastic resonator model~\cite{solin13, solin14} and the linear basis model~\cite{reece14}.

Referring again to the wave and tide data seen in Figure \ref{fig:waveht}, one can see that the peaks and troughs are not perfectly aligned, either because of a delay in one of the sensor readings, or physical delay due to differing sensor locations. It should be possible to model this within the state space approach, allowing for more computationally efficient and interpretable versions of the Gaussian process sensor network model presented in \cite{osborne12}.

\section*{Acknowledgements}
The first author would like to thank T. McClure for the many valuable discussions regarding Kalman filtering.
This work is supported by NIH grant R01-AI107034.

\bibliography{refList}
\bibliographystyle{icml2015}

\end{document}